\documentclass[conference]{IEEEtran}
\IEEEoverridecommandlockouts
\usepackage{cite}
\usepackage{amsmath,amssymb,amsfonts}
\usepackage{algorithmic}
\usepackage{graphicx}
\usepackage{textcomp}
\usepackage{xcolor}
\usepackage{times}
\usepackage{soul}
\usepackage{url}
\usepackage[hidelinks]{hyperref}
\usepackage[utf8]{inputenc}
\usepackage[font=small]{caption}
\usepackage{graphicx}
\usepackage{amsmath}
\usepackage{booktabs}
\usepackage{graphicx}
\usepackage{algorithm}
\usepackage{algorithmic}
\usepackage{tablefootnote}
\usepackage{adjustbox}
\usepackage{tikz}
\usepackage{url}
\usetikzlibrary{arrows,shapes,automata,petri,positioning,calc}
\usepackage{caption}
\tikzset{
    nodeStyle/.style={
        circle,
        thick,
        draw=black,
        fill=white!30,
        minimum size=3mm,
    },
    nodeStyleIncDec/.style={
        circle,
        thick,
        draw=black,
        fill=white!30,
        minimum size=8mm,
    },
    posStyle/.style={
        circle,
        thick,
        draw=black,
        fill=green!30,
        minimum size=8mm,
    },
    negStyle/.style={
        circle,
        thick,
        draw=black,
        fill=red!30,
        minimum size=8mm,
    },
    dottedRectangle/.style={dashed},
}

\def\BibTeX{{\rm B\kern-.05em{\sc i\kern-.025em b}\kern-.08em
    T\kern-.1667em\lower.7ex\hbox{E}\kern-.125emX}}
\usepackage{xcolor, color, soul}
\sethlcolor{white}
\usepackage{float}
\usepackage{placeins}

\begin{document}

\title{Efficient Adversarial Input Generation via\\Neural Net Patching}

\author{\IEEEauthorblockN{Tooba Khan}
\IEEEauthorblockA{
\textit{Indian Institute of Technology Delhi}\\
New Delhi, India \\
www.orcid.org/0000-0002-4001-4242}
\and
\IEEEauthorblockN{Kumar Madhukar}
\IEEEauthorblockA{
\textit{Indian Institute of Technology Delhi}\\
New Delhi, India \\
www.orcid.org/0000-0001-5686-9758}
\and
\IEEEauthorblockN{Subodh Vishnu Sharma}
\IEEEauthorblockA{
\textit{Indian Institute of Technology Delhi}\\
New Delhi, India \\
www.orcid.org/0000-0003-3069-3744}
}

\maketitle

\begin{abstract}

The generation of adversarial inputs has become a crucial issue in establishing the
robustness and trustworthiness of deep neural nets, especially when they are
used in safety-critical application domains such as autonomous vehicles and
precision medicine. However, the problem poses multiple practical challenges, including
scalability issues owing to large-sized networks, and the
generation of adversarial inputs that lack important qualities such as naturalness
and output-impartiality. This problem shares its end goal with the task of
patching neural nets where {\em small} changes in some of the network's weights
need to be discovered so that upon applying these changes, the modified net
produces the desirable output for a given set of inputs. We exploit this
connection by proposing to obtain an adversarial input from a patch, with the
underlying observation that the effect of changing the weights can also be
brought about by changing the inputs instead. Thus, this paper presents a novel way to generate input perturbations that are adversarial for a given network by using an efficient network patching technique.  We note that the
proposed method is significantly more effective than the prior state-of-the-art
techniques.

\end{abstract}

\section*{Introduction}
\label{sec:intro}

Deep Neural Networks (DNNs) today are omnipresent. An important reason behind
their widespread use is their ability to generalize and thereby perform well
even on previously unseen inputs. While this is a great practical advantage, it
may sometimes make DNNs unreliable. In safety- or business-critical
applications this lack of reliability can indeed have dreadful costs.
Evidently, central to a trained network's unreliability lies the lack of
robustness against input perturbations, {\em i.e.}, small changes to some
inputs cause a substantial change in the network's output. This is undesirable
in many application domains. For example, consider a network that has been
trained to issue advisories to aircrafts to alter their paths based on
approaching intruder aircrafts. It is natural to expect such a network to be
robust in its decision-making, i.e. the advisory issued for two very similar
situations should not be vastly different. At the same time, if that is not the
case, then demonstrating the lack of robustness through \emph{adversarial
inputs} can help not only in improving the network but also in deciding when
the network should relinquish control to a more dependable entity.

Given a network and an input, an adversarial input is one that is \emph{very
close} to the given input and yet the network's outputs for the two inputs are
quite different. In the last several years, there has been much work on finding
adversarial
inputs~\cite{whiteBox:1,whiteBox:2,blackBox:1,blackBox:2,blackBox:3}.

These approaches can be divided into black-box and white-box methods based on
whether they consider the network's architecture during the analysis or not. A
variety of techniques have been developed in both these classes, ranging from
generation of random attacks~\cite{blackBox:2,blackBox:3} and gradient-based
methods~\cite{blackBox:1} to symbolic
execution~\cite{symbolic:1,symbolic:2,symbolic:3}, fault
localization~\cite{deepfault}, coverage-guided
testing~\cite{whiteBox:1,whiteBox:2}, and SMT and ILP solving~\cite{smt:1}.
However, there are several issues that limit the practicability of these techniques: a poor success rate, a large distance between the adversarial and the original input (both in terms of the number of input values changed and the degree of the change), unnatural or perceivably different inputs, and output partiality (the techniques’ bias to produce adversarial examples for just one of the network’s outputs). This paper presents a useful approach to generating adversarial inputs in a way that addresses these issues.

We relate the problem of finding adversarial inputs to the task of {\em patching
neural nets}, i.e. applying small changes in some of the network’s weights so
that the modified net behaves desirably for a given set of inputs. Patching
DNNs is a topic of general interest to the Machine Learning community because
of its many applications, which include bug-fixing, watermark resilience, and
fine-tuning of DNNs, among others~\cite{modifications:1}. Intuitively, the
relation between these two problems is based on the observation that a patch
can be translated into an adversarial input because the effect of changing the
weights may be brought about by changing the inputs instead. In fact, a patch
in the very first \emph{edge-layer} of a network can very easily be transformed
into a corresponding change in the input by just solving linear equations.
While there are techniques to solve the patching
problem~\cite{modifications:1,modifications:2}, it is in general a difficult
task, particularly for layers close to the input layer. The computation of the
entire network needs to be encoded and passed to a constraint solver in order
to obtain a patch. For large-sized networks, this gives rise to a big
monolithic constraint leading to scalability issues for the solver. We address
this by proposing an improvement in the technique of~\cite{modifications:1},
and then using it repeatedly to find a middle-layer patch and chop off the
latter half of the network, till a first-layer patch has been obtained. 
\hl{Our improved patching technique not only gives us a smaller patch, but when used with our adversarial image generation technique, it also helps in the generation of more natural adversarial inputs.}
Our
experiments on three popular image-dataset benchmarks show that our approach
does significantly better than other state-of-the-art techniques, in terms of
\hl{the naturalness and} the number of pixels modified as well as the magnitude of the change.  This
reflects in the quality of the adversarial images, both visually and in several
qualitative metrics that we present later.

The core contributions of this paper are:
\begin{itemize}
	\item A novel approach of producing perturbations of inputs which are
		adversarial for a given network, with the help of an
		\emph{efficient patching technique}\footnote{Though not our
		primary contribution, our patching technique is an improvement
		over~\cite{modifications:1} (see
		Sect.~\ref{subsec:modimprove}).}.

	\item An extensive experimental evaluation using CIFAR-10~\cite{cifar},
		MNIST~\cite{mnist}, and ImageNet~\cite{imagenet} datasets, and
		a number of qualitative parameters, to show the efficacy of our
		approach over the state-of-the-art.
\end{itemize}

\section*{Related Work}
\label{sec:related}

The robustness of DNNs has gained much attention in the last several years as DNNs are permeating our lives, even safety-and business-critical aspects. A number of techniques have been developed to establish robustness or demonstrate the lack of it through adversarial examples. These techniques are broadly classified as black-box, gradient-based and white-box approaches.

Black box methods do not consider the architecture of DNNs in trying to argue about their robustness. Attacks based on the L2 and L0 norms were introduced in \cite{blackBox:1}. These attacks change the pixel values by some fixed amount and measure their effectiveness by the decrease in confidence of the original label. 

Fast gradient sign method~\cite{blackBox:2} is a gradient-based method, which uses gradient of the neural network when the original and modified images are fed to it. Output diversification \cite{ods} is another gradient-based technique that aims at maximizing the output diversity, which we measure using the Pielou score\footnote{refer to Metrics of Evaluation (section \ref{sec:Metrics} in Experimental Setup) }. These methods modify many pixels in the original image to induce a misclassification which makes the adversarial image visibly different from the original image.

White box methods, on the other hand, involve looking at the complete architecture of the DNNs to obtain adversarial inputs or argue that none exists. Verification of neural networks using Symbolic execution is one such white box approach. As discussed in \cite{smt:1}, it translates the neural network into an imperative program and uses SMT (Symbolic Modulo Theory) based solver to prove given properties. However, such techniques are not scalable due to exponential time complexity. Symbolic propagation, as discussed in \cite{symbolic:3}, \cite{symbolic:1} and \cite{symbolic:2} converts inputs to symbolic representations and propagates them through the hidden and output layers. But, these techniques often give loose bounds and lack precision. 

Another white box technique is to find flaws in the training phase of the neural network, such as the use of an inappropriate loss function~\cite{lossFunction:1,lossFunction:2}. 
But, such techniques are still vulnerable to adversarial attacks and can not be transferred or used to test the robustness of existing DNNs.
DeepFault~\cite{deepfault} uses fault localization, i.e., finding the areas of the network that are mainly responsible for incorrect behaviors. 
Coverage-based white box techniques use structural coverage metrics such as neuron coverage and modified condition/decision coverage (MC/DC). \cite{whiteBox:3} developed a tool that uses MC/DC variants for verification of neural networks using neuron coverage. \cite{fuzz:3,fuzz:2} use mutation-based  and genetic-algorithm based strategies to generate test cases that can maximize neuron coverage.
\cite{whiteBox:1,whiteBox:2} implements a white-box approach that maximizes neuron coverage and differential behavior of a group of DNNs.

Even though the code coverage criteria of software engineering test methodologies correspond to neuron coverage, it is not a useful indicator of the production of adversarial inputs. According to authors in \cite{neuronCoverage}, Neuron coverage statistics lead to the detection of fewer flaws, making them inappropriate for proving the robustness of DNNs. They also present three new standards -- defect detection, naturalness, and output impartiality -- that can be used to gauge the quality of adversarial inputs produced as alternatives to the L2 and L-$\infty$ norms. Their findings establish that the adversarial image set generated using neuron coverage measures did not perform well on these three standards. 

Since our technique relies on finding modifications in DNNs, we also discuss recent work in this respect. \cite{modifications:1} proposes a technique to find minimal modifications in a single layer in a DNN to meet a given outcome. This work has been extended further in~\cite{modifications:2} to come up with multi-layer modifications by dividing the problem into multiple subproblems and applying the idea of~\cite{modifications:1} on each one of them. Our work proposes an improvement over their idea and uses the improved technique to find small modifications, which are then translated into adversarial inputs.

\section*{Illustrative Example}
\label{sec:example}

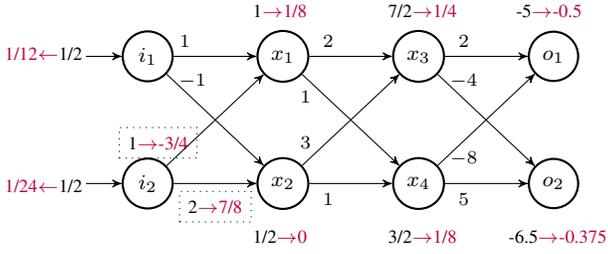
\begin{figure}[ht]
	\centering
	\begin{tikzpicture}[node distance=1cm and 1.1cm,>=stealth',auto,initial text={}, every place/.style={draw}]
        \node[draw,align=left,dotted] at (0.14,-1.15) {\scriptsize 1\textcolor{purple}{$\rightarrow$-3/4}};
        \node[draw,align=left,dotted] at (0.9,-2.0) {\scriptsize 2\textcolor{purple}{$\rightarrow$7/8}};
        
        \node[text width=3cm] at (-0.4,0.03) {\scriptsize \textcolor{purple}{1/12$\leftarrow$}1/2};
        \node[text width=3cm] at (-0.4,-1.72) {\scriptsize \textcolor{purple}{1/24$\leftarrow$}1/2};
        
        \node[text width=3cm] at (2.9,-2.4) {\scriptsize 1/2\textcolor{purple}{$\rightarrow$0}};
        \node[text width=3cm] at (4.7,-2.4) {\scriptsize 3/2\textcolor{purple}{$\rightarrow$1/8}};
        \node[text width=3cm] at (6.3,-2.4) {\scriptsize -6.5\textcolor{purple}{$\rightarrow$-0.375}};
         \node[text width=3cm] at (2.9,0.6) {\scriptsize 1\textcolor{purple}{$\rightarrow$1/8}};
         \node[text width=3cm] at (4.7,0.6) {\scriptsize 7/2\textcolor{purple}{$\rightarrow$1/4}};
         \node[text width=3cm] at (6.4,0.6) {\scriptsize -5\textcolor{purple}{$\rightarrow$-0.5}};

		\node [nodeStyle,initial] (i1) {\footnotesize $i_{1}$};
		\node [nodeStyle,initial] (i2) [below=of i1] {\footnotesize $i_{2}$};
		\node [nodeStyle] (x1) [right=of i1] {\footnotesize $x_{1}$};
		\node [nodeStyle] (x2) [right=of i2] {\footnotesize $x_{2}$};
		\node [nodeStyle] (x3) [right=of x1] {\footnotesize $x_{3}$};
		\node [nodeStyle] (x4) [right=of x2] {\footnotesize $x_{4}$};
		\node [nodeStyle] (o1) [right=of x3] {\footnotesize $o_{1}$};
		\node [nodeStyle] (o2) [right=of x4] {\footnotesize $o_{2}$};
		
		\path[->] (i1) edge node[above, xshift=-0.4cm, yshift=0cm] {\scriptsize $1$} (x1);
		\path[->] (i2) edge node[below, yshift=0cm] {} (x2);
		\path[->] (i1) edge node[above, xshift=-0.3cm, yshift=0.3cm] {\scriptsize $-1$} (x2);
		\path[->] (i2) edge node[above, xshift=-1.3cm, yshift=-0.5cm] {} (x1);
		\path[->] (x1) edge node[above, xshift=-0.3cm, yshift=0cm] {\scriptsize $2$} (x3);
		\path[->] (x2) edge node[below, xshift=-0.3cm, yshift=0cm] {\scriptsize $1$} (x4);
		\path[->] (x2) edge node[above, xshift=-0.6cm, yshift=-0.5cm] {\scriptsize $3$} (x3);
		\path[->] (x1) edge node[above, xshift=-0.6cm, yshift=0.1cm] {\scriptsize $1$} (x4);
		\path[->] (x3) edge node[above, xshift=-0.3cm, yshift=0cm] {\scriptsize $2$} (o1);
		\path[->] (x4) edge node[below, xshift=-0.3cm, yshift=0cm] {\scriptsize $5$} (o2);
		\path[->] (x3) edge node[above, xshift=-0.3cm, yshift=0.3cm] {\scriptsize $-4$} (o2);
		\path[->] (x4) edge node[below, xshift=-0.3cm, yshift=-0.3cm] {\scriptsize $-8$} (o1);

	\end{tikzpicture}
	\caption{Adversarial inputs from first-layer modification. The adversarial input and the corresponding values of each neuron are written in red. The modification required in first layer weights are shown in dotted boxes.}
	\label{fig1}
\end{figure}

Let us consider a toy DNN, $\mathcal{N}$, shown in Fig.~\ref{fig1}. It has two
neurons in each of its four layers -- the input layer, followed by two
hidden layers, and then the output layer. We assume that the hidden layers have
ReLU\footnote{ReLU($x$) = $\mathit{max}(0,x)$} as the activation function. For
neurons without an activation function, the value of a neuron is computed by
summing up, for each incoming edge to the neuron, the product of the edge
weight and the value of the neuron at the edge's source. In presence of an
activation function, the value is computed by applying the function on this
sum. For example, for the input $\langle 0.5,0.5 \rangle$, the values at the
next three layers are $\langle 1,0.5 \rangle$, $\langle 3.5,1.5 \rangle$, and
$\langle -5,-6.5 \rangle$ respectively.

Finding an adversarial input $\langle i_1,i_2 \rangle$ for the input $\langle
0.5,0.5 \rangle$ amounts to finding a value for each $i_k$ ($k \in \{1,2\}$)
such that $|i_k-0.5| \leq \delta$ and the corresponding output $o_2 > o_1$, for
a given small $\delta$. It is noteworthy that if we want the second output to
become bigger than the first one, this can be achieved by modifying the weights
instead of the inputs. For example, if the edges connecting the second input
neuron to the first hidden layer had the weights $\langle -0.75,0.875 \rangle$
instead of $\langle 1,2 \rangle$, then the next three layers would have the
values $\langle 0.125,0 \rangle$, $\langle 0.25,0.125 \rangle$, and $\langle
-0.5,-0.375 \rangle$, and our goal would be met by modifying the weights while
keeping the inputs unchanged. We will come to the question of how to find the
changed first-layer weights in a bit, but let us first see how the changed
weights can help us obtain an adversarial input. This is a rather simple
exercise. Note that with the changed weights, the values of the neurons in the
first hidden layer were $\langle 0.125,0 \rangle$. So, our task is simply to
find inputs for which the first hidden layer values stay as $\langle 0.125,0
\rangle$, but with the original weights $\langle 1,2 \rangle$. This can be done
by solving the following equations, where $\delta_1$ and $\delta_2$
($\delta_1,\delta_2$ $\leq \delta \leq 0.5$, say) are the changes in the two
inputs respectively.

\begin{equation}
\label{eq1}
    (1/2+\delta_1) + (1/2+\delta_2) = 1/8
\end{equation}
\begin{equation}
\label{eq2}
    -(1/2+\delta_1) + 2*(1/2+\delta_2) =0
\end{equation}

We get $\delta_1=-5/12, \delta_2=-11/24$ and, thus, the adversarial input as
$\langle 1/12,1/24 \rangle$. The dotted rectangles in Fig.~\ref{fig1} contain
the adversarial inputs and the corresponding values at each layer.

The first thing to notice here is that Eqn.~\ref{eq2} could have been relaxed
as $-(1/2+\delta_1) + 2*(1/2+\delta_2) \leq 0$ because of the ReLU activation
function. This may give us smaller values of $\delta_i$'s. Moreover, along with
minimizing the change in each input pixel, we can also minimize the number of
pixels that are modified, as shown here.
\begin{equation}
    (1/2+\delta_1*M_1) + (1/2+\delta_2*M_2) = 1/8
\end{equation}
\begin{equation}
    -(1/2+\delta_1*M_1) + 2*(1/2+\delta_2*M_2) \leq 0
\end{equation}
\begin{equation}
	M_1,M_2 \in \{0,1\};~~~ \mathit{minimize}~~\sum M_i
\end{equation}

In pratice, we solve these constraints in place of Eqns.~\ref{eq1}-\ref{eq2};
this gives us better adversarial inputs.

There are a few more points to note before we proceed. First, it is only the
first-layer modification that may be easily translated into an adversarial
input as illustrated. Modification in deeper layers are not immediately
helpful; they cannot be translated easily into an adversarial input because of
the non-linear activation functions. Second, we need to find \emph{small}
modifications in the weights, so that the corresponding $\delta_i$'s in the
inputs fall within the allowed $\delta$. And, lastly, while there are ways to
compute a first-layer change directly using an SMT or an ILP
solver~\cite{modifications:1}, this approach is not very scalable in practice
as large-sized networks give rise to big monolithic formulas that may be
difficult for the solver. Instead, we propose an iterative approach that finds
a middle-layer modification using \hl{an improved version of} ~\cite{modifications:1}\footnote{\hl{We have described the improved version in the next section. We also show the improvement quantitatively in the results section.}}
 and
chops off the latter half of the network, repeatedly till a first-layer patch
is found.

Since our network has three edge-layers, we start by finding a small
modification of the weights in the second edge-layer with which we can achieve
our target of making $o_2$ bigger than $o_1$ for the input $\langle 0.5,0.5 \rangle$. We
propose an $\epsilon_{i,j}$ change in the weight of the $j^{th}$ edge in
$i^{th}$ edge-layer, and encode the constraints for $o_2 > o_1$ as follows:

\begin{equation}
\label{eq3}
	x_3 = \mathit{max}(0, 1*(2+\epsilon_{2,1}) + 1/2*(3+\epsilon_{2,2}))
\end{equation}
\begin{equation}
\label{eq4}
	x_4 = \mathit{max}(0, 1*(1+\epsilon_{2,3}) + 1/2*(1+\epsilon_{2,4}))
\end{equation}
\begin{equation}
\label{eq5}
    -4*x_3 + 5*x_4 > 2*x_3 - 8*x_4
\end{equation}

\begin{figure}[ht]
	\centering
	\begin{tikzpicture}[node distance=1cm and 1.1cm,>=stealth',auto,initial text={\scriptsize $1/2$}, every place/.style={draw}]
        
        \node[rectangle, dottedRectangle,
        draw = red,
        text = red,
        minimum width = 5.3cm, 
        align=left,
        minimum height = 4cm] (r) at (1.9,-0.6) {};
        \node[text width=3cm] at (0.9,1.2) {\textcolor{red}{\scriptsize Extracted Network}};
        
        \node[text width=3cm] at (3.2,-2.4) {\scriptsize 1/2};
        \node[text width=3cm] at (4.7,-2.4) {\scriptsize 3/2\textcolor{purple}{$\rightarrow$1/8}};
        \node[text width=3cm] at (6.3,-2.4) {\scriptsize -6.5\textcolor{purple}{$\rightarrow$-0.375}};
        \node[text width=3cm] at (3.25,0.6) {\scriptsize 1};
        \node[text width=3cm] at (4.7,0.6) {\scriptsize 7/2\textcolor{purple}{$\rightarrow$1/4}};
        \node[text width=3cm] at (6.5,0.6) {\scriptsize -5\textcolor{purple}{$\rightarrow$-0.5}};

		\node [nodeStyle,initial] (i1) {\footnotesize $i_{1}$};
		\node [nodeStyle,initial] (i2) [below=of i1] {\footnotesize $i_{2}$};
		\node [nodeStyle] (x1) [right=of i1] {\footnotesize $x_{1}$};
		\node [nodeStyle] (x2) [right=of i2] {\footnotesize $x_{2}$};
		\node [nodeStyle] (x3) [right=of x1] {\footnotesize $x_{3}$};
		\node [nodeStyle] (x4) [right=of x2] {\footnotesize $x_{4}$};
		\node [nodeStyle] (o1) [right=of x3] {\footnotesize $o_{1}$};
		\node [nodeStyle] (o2) [right=of x4] {\footnotesize $o_{2}$};
		
		\path[->] (i1) edge node[above, xshift=-0.3cm, yshift=0cm] {\scriptsize $1$} (x1);
		\path[->] (i2) edge node[below, xshift=-0.2cm, yshift=0cm] {\scriptsize $2$} (x2);
		\path[->] (i1) edge node[below, xshift=-0.7cm, yshift=0.6cm] {\scriptsize $-1$} (x2);
		\path[->] (i2) edge node[above, xshift=-0.6cm, yshift=-0.5cm] {\scriptsize $1$} (x1);
		\path[->] (x1) edge node[above, xshift=0cm, yshift=0cm] {\scriptsize 2\textcolor{purple}{$\rightarrow$7/8}} (x3);
		\path[->] (x2) edge node[below, xshift=0cm, yshift=0cm] {\scriptsize 1\textcolor{purple}{$\rightarrow$3/4}} (x4);
		\path[->] (x2) edge node[above, xshift=-0.28cm, yshift=-0.65cm] {\scriptsize 3\textcolor{purple}{$\rightarrow$-5/4}} (x3);
		\path[->] (x1) edge node[above, xshift=-0.27cm, yshift=0.24cm] {\scriptsize 1\textcolor{purple}{$\rightarrow$-1/4}} (x4);
		\path[->] (x3) edge node[above, xshift=-0.3cm, yshift=0cm] {\scriptsize $2$} (o1);
		\path[->] (x4) edge node[below, xshift=-0.3cm, yshift=0cm] {\scriptsize $5$} (o2);
		\path[->] (x3) edge node[above, xshift=-0.3cm, yshift=0.3cm] {\scriptsize $-4$} (o2);
		\path[->] (x4) edge node[below, xshift=-0.3cm, yshift=-0.3cm] {\scriptsize $-8$} (o1);
	\end{tikzpicture}
	\caption{Middle-layer modification and sub-net extraction}
	\label{fig2}
\end{figure}
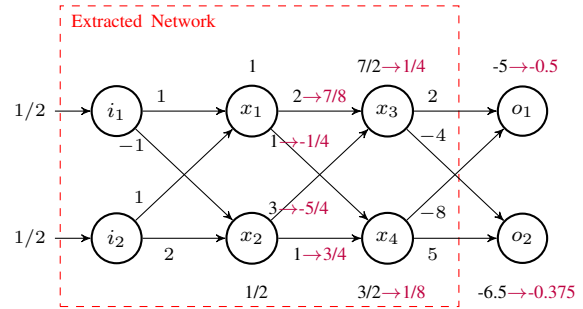

The range of each $\epsilon_{i,j}$ is [$-\alpha, \alpha$] if $\alpha$ is the
biggest permissible modification for an edge-weight. We minimize the
magnitude of the total change using Gurobi~\cite{gurobi}. For the equations
above, we get
$\langle \epsilon_{2,1},\epsilon_{2,2},\epsilon_{2,3},\epsilon_{2,4} \rangle = \langle -9/8,-17/4,-5/4,-1/4 \rangle$.
These changes indeed make the second output bigger (see Fig.~\ref{fig2}).

Our next step is to extract a sub-network (see Fig.~\ref{fig2}) and look for a
modification in its middle layer. Since the extracted network has only two
edge-layers, this step will give us a first-layer modification. The
equations, subject to the constraint that $x_3$ and $x_4$ get the values 1/4
and 1/8 respectively, are shown below.
\begin{equation}
\label{eq6}
	x_1 = \mathit{max}(0, 1/2*(1+\epsilon_{1,1}) + 1/2*(1+\epsilon_{1,2}))
\end{equation}
\begin{equation}
\label{eq7}
	x_2 = \mathit{max}(0, 1/2*(-1+\epsilon_{1,3}) + 1/2*(2+\epsilon_{1,4}))
\end{equation}
\begin{equation}
\label{eq8}
    2*x_1 + 3*x_2 = 1/4; ~~~ 1*x_1 + 1*x_2 = 1/8
\end{equation}

Gurobi gives us the solution $\langle
\epsilon_{1,1},\epsilon_{1,2},\epsilon_{1,3},\epsilon_{1,4} \rangle = \langle
0,-7/4,0,-9/8 \rangle$, from which we can obtain the adversarial input $\langle
1/12,1/24 \rangle$ as already shown above.

\section*{Methodology}
\label{sec:method}

We now describe the technical details of our approach and present our
algorithm. We begin with the notation that we use. Let $\mathcal{N}$ denote the
DNN\footnote{Our approach is not limited to DNNs. It can be easily extended to CNNs(Convoluted Neural networks) where the modification is found in the first fully connected layer and then translated back to the input layer. The convolutional and pooling layers do not limit our methodology.} that we have with $n$ inputs ($i_1, i_2,\ldots,i_n$), $m$ outputs ($o_1,
o_2,\ldots, o_m$), and $k$ layers ($l_1, l_2,\ldots, l_k$). We use $v_{p,q}$ to
denote the value of the $q^{th}$ neuron in $l_p$.  We assume that $\mathcal{N}$
is feed-forward, i.e., the (weighted) edges connect neurons in adjacent layers
only, and point in the direction of the output layer.  We use the term
\emph{edge-layer} to refer to all the edges between any two adjacent layers of
$\mathcal{N}$, and denote the edge layers as $el_1, el_2,\ldots, el_{k-1}$.  We
also assume the hidden layers ($l_2, l_3,\ldots, l_{k-1}$) in $\mathcal{N}$
have ReLU activation function, and that there is no activation function on any
output neuron. For simplicity, we assume that the neurons do not have any
biases. This is not a limitation in any sense; our implementation handles them
directly. Moreover, a DNN with biases can be converted into an equivalent one
without any biases. Like in the previous section, we use $\delta_p$ to denote
the change in the $p^{th}$ input, and $\epsilon_{q,s}$ to denote the change in
the weight of the $q^{th}$ edge in $el_s$. The $\delta_p$'s are constrained to
be $\leq \delta$, which is the biggest perturbation allowed in any pixel to
find an adversarial input.

Algorithm~\ref{alg:alg1} presents a pseudocode of our algorithm. The inputs to
the algorithm are: $\mathcal{N}, l, \delta$, and the values for the input neurons
$v_{1,1}, v_{1,2},\ldots,v_{1,n}$, for which the corresponding output does not
satisfy a given adversarial property $\phi(o_1,o_2,\ldots, o_m)$ (denoted
simply as $\phi$, henceforth). The aim of our algorithm is to find a new set of
input values $v'_{1,1}, v'_{1,2},\ldots,v'_{1,n}$ such that $\mathcal{N}$'s
output corresponding to these new inputs satisfies $\phi$. The algorithm works
in the following two steps. First, it finds a \emph{small} modification in the
weights of $el_1$ to derive $\mathcal{N}_{mod}$ (which is essentially
$\mathcal{N}$ with the modified weights in $el_1$), such that the output of
$\mathcal{N}_{mod}$ for the input $v_{1,1}, v_{1,2},\ldots,v_{1,n}$ satisfies
$\phi$. Then, the algorithm translates this first-layer modification into
adversarial inputs $v'_{1,1}, v'_{1,2},\ldots,v'_{1,n}$, subject to the
constraint that $|v'_{1,p}-v_{1,p}| \leq \delta$, for every $p \in [1,n]$. This
second step is shown in the algorithm as the function $\mathit{mod2adv}$, the
pseudocode of which has been omitted as this is a simple call to Gurobi as
illustrated in the previous section.

Let us assume for the time being that we have a sub-routine
$\mathit{modifyEdgeLayer}$ that takes as input a DNN $\mathcal{N}$, one of its
edge-layers $el_k$, values of the input neurons $v_{1,1},
v_{1,2},\ldots,v_{1,n}$, and a property $\phi$ on the output layer neurons, and
returns a new network $\mathcal{N}_{mod}$ with the constraints that:

\begin{itemize}
	\item $\mathcal{N}_{mod}$ is same as $\mathcal{N}$ except for the weights in $el_k$, and
	\item the output of $\mathcal{N}_{mod}$ on $v_{1,1}, v_{1,2},\ldots,v_{1,n}$ satisfies $\phi$.
\end{itemize}

Clearly, with such a sub-routine, the first step of our algorithm becomes
trivial. We would simply call $\mathit{modifyEdgeLayer}$ with the given input,
$\mathcal{N}$, $\phi$, and $el_1$. We refer to the work of~\cite{modifications:1} which gives us exactly this. However, we do not use
it directly to find our first-layer modification. Informally, the technique
of~\cite{modifications:1} uses variables $\epsilon_{q,s}$ to denote the changes
in the weights (in a given edge-layer $s$) and encodes the computation of the
entire network, and then adds the constraint that the output must satisfy
$\phi$. It then uses \hl{Marabou}~\cite{marabou} on these constraints, to solve for (and optimize)
the values of $\epsilon_{q,s}$. Consider an example (reproduced
from~\cite{modifications:1}) shown in Fig.~\ref{fig3} with the output property
$\phi := (v_{3,1} \geq v_{3,2})$. Let us ignore the color of the output neurons
for now. If the input neurons are given values $v_{1,1}=3$ and $v_{1,2}=4$, the
output neurons get the value $v_{3,1}=-2$ and $v_{3,2}=2$, which does not
satisfy $\phi$. Note that the hidden layer neurons have ReLU activation
function. In order to obtain a second edge-layer modification such that $\phi$
holds for the input $\langle 3,4 \rangle$, the technique
of~\cite{modifications:1} generates the constraints given in Fig.~\ref{fig3cons}.

\begin{figure}[ht]
	\centering
	\begin{tikzpicture}[node distance=1cm and 2.1cm,>=stealth',auto,initial text={}, every place/.style={draw}]
        
		\node [nodeStyleIncDec,initial] (i1) {\footnotesize $v_{1,1}$};
		\node [nodeStyleIncDec,initial] (i2) [below=of i1] {\footnotesize $v_{1,2}$};
		\node [nodeStyleIncDec] (x1) [right=of i1] {\footnotesize $v_{2,1}$};
		\node [nodeStyleIncDec] (x2) [right=of i2] {\footnotesize $v_{2,2}$};
		\node [posStyle] (o1) [right=of x1] {\footnotesize $v_{3,1}$};
		\node [negStyle] (o2) [right=of x2] {\footnotesize $v_{3,2}$};
		
		\path[->] (i1) edge node[above, xshift=-0.3cm, yshift=0cm] {1} (x1);
		\path[->] (i2) edge node[below, xshift=-0.2cm, yshift=0cm] {-1} (x2);
		\path[->] (i1) edge node[below, xshift=-0.9cm, yshift=0.6cm] {2} (x2);
		\path[->] (i2) edge node[above, xshift=-0.8cm, yshift=-0.5cm] {-2} (x1);
		
		\path[->] [text width=2cm, align=center] (x1) edge node[above, xshift=0cm, yshift=0cm] {1} (o1);
		\path[->] (x2) [text width=2cm, align=center] edge node[below, xshift=0cm, yshift=0cm] {1} (o2);
		\path[->] (x1) [text width=2cm, align=center] edge node[below, xshift=-1.0cm, yshift=0.6cm] {-1} (o2);
		\path[->] (x2) edge node[above, xshift=-0.8cm, yshift=-0.5cm] {-1} (o1);
		
		
	\end{tikzpicture}
	\caption{Example illustrating DNN modification, from \protect\cite{modifications:1}. \hl{Red indicated decrement neuron and green indicates increment neuron.}}
	\label{fig3}
\end{figure}
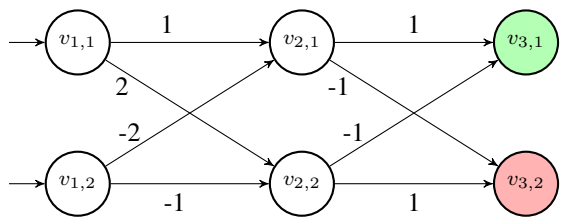

\begin{figure}
\begin{equation}
	\mathit{minimize} ~~ M
\end{equation}
\begin{equation}
	M \geq 0
\end{equation}
\begin{equation}
	-M \leq \epsilon_{1,2} \leq M
\end{equation}
\begin{equation}
	-M \leq \epsilon_{2,2} \leq M
\end{equation}
\begin{equation}
	-M \leq \epsilon_{3,2} \leq M
\end{equation}
\begin{equation}
	-M \leq \epsilon_{4,2} \leq M
\end{equation}
\begin{equation}
	v_{3,1} = 0.(1+\epsilon_{1,2}) + 2.(-1+\epsilon_{2,2})
\end{equation}
\begin{equation}
	v_{3,2} = 0.(-1+\epsilon_{3,2}) + 2.(1+\epsilon_{4,2})
\end{equation}
\begin{equation}
	v_{3,1} \geq v_{3,2}
\end{equation}
	\caption{DNN modification constraints for Fig.~\ref{fig3}}
	\label{fig3cons}
\end{figure}

In a similar way, the constraints can be generated for modification in any
layer, by propagating the input values up to that layer, encoding the
computation from there onward, and adding the output property $\phi$. If a
first-layer modification has to be found, this gives rise to a big monolithic
constraint, particularly for large-sized networks. This does not scale in
practice, and therefore we propose an iterative approach of doing this in our
algorithm. The iterative approach uses the above technique to find a
middle-layer modification, derive a new output property $\phi'$ for the first
half of the network, and does this repeatedly till a first-layer modification
is found.  This has been illustrated in the function
$\mathit{findFirstLayerMod}$ in Alg.~\ref{alg:alg1}, which calls
$\mathit{modifyEdgeLayer}$ in a loop. The last bit here is to understand how
the modified output property $\phi'$ may be derived in each iteration. The way
this works is as follows. Let us say that the last call to
$\mathit{modifyEdgeLayer}$ was made on edge-layer $el_{(j-1)}$, which connects
the layers $l_{(j-1)}$ and $l_j$. We can use the modified weights to propagate
the inputs all the way to layer $l_j$, by simply simulating the network on the
inputs. This gives us values for all the neurons in layer $l_j$, say $c_1, c_2,
\ldots$ and so on. Now, the modified weights are replaced by the original
weights, and all the layers after $l_j$ are dropped off from the network.  This
reduced network $\mathcal{N}'$ has exactly the layers $l_1$ to $l_j$ of
$\mathcal{N}$. We denote this reduction as $\mathcal{N}\downarrow_{(l_1 \ldots
l_j)}$. Ideally, we would want to find a middle-layer modification in
$\mathcal{N}'$ under the new output constraint as $\phi' := (v_{j,1} = c_1)
\wedge (v_{j,2} = c_2) \wedge \ldots$ and so on. The updated $\phi'$ is correct
because we know that $\phi$ gets satisfied when these values are propagated
further to the output layer. But, we can relax the strict equalities of $\phi'$
into inequalities as discussed in the next subsection.

\begin{algorithm}[ht]
\caption{Adversarial Inputs via Network Patching}
\label{alg:alg1}
	\textbf{Input}: $\mathcal{N}, l, \delta, \phi,$ and input $\langle v_{1,1},\ldots,v_{1,n} \rangle$
	\par
	\textbf{Output}: Adversarial input $\langle v'_{1,1},\ldots,v'_{1,n} \rangle$
	\hfill\\

\textbf{findFirstLayerMod($\mathcal{N}, l, \mathit{in}, \phi$):} 
\par
\begin{algorithmic}[1]
	\WHILE{true}
	\STATE $p \gets \lceil (l/2) \rceil$
	\STATE $\mathcal{N}_{mod} \gets \mathit{modifyEdgeLayer}(\mathcal{N}, \mathit{in}, \phi, el_p)$
	\STATE \textbf{return} $\mathcal{N}_{mod}$ \textbf{if}$(p = 1)$

	\STATE $\langle c_1,c_2,\ldots \rangle \gets \mathit{simulate}(\mathcal{N}_{mod}, \mathit{in})$
	\STATE $\phi' = (v_{(p+1),1}=c_1) \wedge (v_{(p+1),2}=c_2) \wedge \ldots$
	\STATE $\mathcal{N}' = \mathcal{N}\downarrow_{(l_1 \ldots l_{(p+1)})}$
	\STATE $\mathcal{N} \gets \mathcal{N}'$; $l \gets (p+1)$; $\phi \gets \phi'$
	\ENDWHILE
\end{algorithmic}
\hfill\\
\textbf{main():}
\begin{algorithmic}[1] 
	\STATE $\mathit{in} \gets \langle v_{1,1},\ldots,v_{1,n} \rangle$
	\STATE $\mathcal{N}_{mod} \gets \mathit{findFirstLayerMod}$($\mathcal{N}, l, \mathit{in}, \phi$)
	\STATE $\langle \delta_1, \delta_2,\ldots,\delta_n \rangle$ = $\mathit{mod2adv}$($\mathcal{N}_{mod}, \mathit{in}, \delta$)
    \STATE $\langle v'_{1,1},\ldots,v'_{1,n} \rangle$ = $\langle v_{1,1},\ldots,v_{1,n} \rangle$ + $\langle \delta_1,\ldots,\delta_n \rangle$
    \STATE \textbf{return $\langle v'_{1,1},\ldots,v'_{1,n} \rangle$}
\end{algorithmic}
\end{algorithm}


\subsection{Simplifying DNN Modification Constraints}
\label{subsec:modimprove}

Let us revisit the example of Fig.~\ref{fig3}, and the constraints
corresponding to the modification problem shown in Fig.~\ref{fig3cons}.
Recollect that the modification problem in this example was to find small
changes in the weights of the second edge-layer, such that $v_{3,1} \geq
v_{3,2}$ for the input $\langle 3,4 \rangle$. This is not true for the DNN in
this example as $v_{3,1}$ gets the value -2, whereas $v_{3,2}$ gets the value
2. A possible way of satisfying $v_{3,1} \geq v_{3,2}$ is to change weights in
such a way that $v_{3,2}$ \emph{decreases} and $v_{3,1}$ \emph{increases}. This
can give us a marking of the final layer neurons as decrement and increment,
which has been indicated by the neuron colors red and green in Fig.~\ref{fig3}.
The useful thing about such a marking is that it can be propagated backward to
other layers. For instance, in the same example, $v_{2,1}$ and $v_{2,2}$ can
also be colored green and red, resp. This works by looking at the edge weights.
Since $v_{2,1}$ is connected to $v_{3,1}$ with a positive-weight edge, an
increase in $v_{3,1}$ can be brought about by \emph{increasing} $v_{2,1}$. If
we look at $v_{2,2}$, since it connected to $v_{3,1}$ with a negative-weight
edge, a \emph{decrease} in $v_{2,2}$ would result in an increase in $v_{3,1}$.
This marking was proposed by Elboher et al.~\cite{DBLP:conf/cav/ElboherGK20},
although it was in the context of abstraction-refinement of neural networks. We
refer the interested readers to~\cite{DBLP:conf/cav/ElboherGK20} for more
details about this marking scheme. In what follows, we explain how this marking
can be useful in simplifying the modification constraints.

Since we are interested in modifying weights in the second edge-layer ($el_2$),
we propagate the inputs to the second layer of neurons. This gives us the
values $\langle 0,2 \rangle$ for $\langle v_{2,1},v_{2,2} \rangle$. Having
propagated the input to the source neurons of $el_2$, and the
increment-decrement marking at the target neurons of $el_2$, we claim that we
can identify whether a given edge-weight should be increased, or decreased. Let
us consider the edge between $v_{2,2}$, which has a value of 2, and $v_{3,1}$,
which has an increment marking. We claim that the change in this edge,
$\epsilon_{2,2}$ should be positive. Naturally, since the value is positive, we
should multiply with a \emph{bigger} weight to get an increased output.
Instead, if a positive value was connected to a decrement neuron, we should
decrease the weight (for example, for the edge between $v_{2,2}$ and
$v_{3,2}$). In case of negative values, just the opposite needs to be done. And
if the value is zero, no change needs to be made at all. With this, the
constraints in Fig.~\ref{fig3cons} get simplified as $\epsilon_{1,2} =
\epsilon_{2,2} = 0, 0 \leq \epsilon_{3,2} \leq M$ and $-M \leq \epsilon_{3,2}
\leq 0$. We have implemented\footnote{As compared to~\cite{modifications:1} on
their benchmarks, our implementation produces smaller modifications, and does
it an order of magnitude faster on an average. Refer to section \ref{sec:results} for our comparisons with \cite{modifications:1}.} this on top of the tool
corresponding to~\cite{modifications:1} and used it in our call to
$\mathit{modifyEdgeLayer}$.

We end this section with a brief note on how this increment decrement marking
may help us relax the modified output property $\phi'$. Recollect that $\phi'$
was derived as a conjunction of equality constraints, where each conjunct was
equating a last-layer neuron of the reduced network with the values obtained by
simulating the input on $\mathcal{N}_{mod}$. Since we know the
increment-decrement marking of last layer neurons, we can relax each conjunct
into an inequality by replacing the equality sign with $\leq$ ($\geq$) for
decrement (increment) neurons. This allows us to obtain, possibly better,
solutions more often.

\section*{Experimental Setup}
\label{sec:setup}

We implemented our approach in a tool called \textsc{Aigent\footnote{https://github.com/KhanTooba/AIGENT.git}} ({\bf A}dversarial
{\bf I}nput {\bf Gen}era{\bf t}or), using the Tensorflow and Keras libraries
for working with the DNNs. \textsc{Aigent} uses Gurobi for constrained
optimization. We ran \textsc{Aigent} with fully connected deep neural networks which have 5-10 hidden layers where each layer has neurons ranging from 10 to 50. The results we present in Table:~\ref{tab:2} are average results obtained from all the experiments performed.

\subsection{Benchmark datasets}
We conducted our experiments on three popular datasets: MNIST, CIFAR-10, and
ImageNet. We chose these benchmarks because they are readily available and are
acceptable as inputs by a number of tools, making it easier to compare
different techniques in a fair way. MNIST consists of 60,000 black and white
images of handwritten digits for the purpose of training and 10,000 for
testing. Each image is of 28x28 size. It has 10 classes with labels
corresponding to each digit. CIFAR-10 consists of 60000 32x32 colour images in
10 classes. ImageNet is a large dataset which consists of images in 1000
classes. 

\subsection{Metrics of Evaluation}
\label{sec:Metrics}
In addition to the usual metrics like L2 and L-$\infty$ distance, and the time
taken, we have used the following metrics to compare our results with the
results of existing approaches.

\begin{enumerate}
    \item Defect detection: 
    \begin{equation*}
	    \frac{\mathit{Number~of~images~successfully~attacked}}{\mathit{Total~number~of~images}}*100
    \end{equation*}
   A low defect detection implies that the method could generate adversarial images for a limited number of the original images and hence it is undesirable.

   We conducted experiments on a set of 10,000 MNIST images. \textsc{Aigent} (high defect) was able to produce adversarial images for 9140 original images. Thus, the defect detection rate for this case is 91.4\% ($2^{nd}$ row under MNIST in Table \ref{tab:2}).
	    
    \item Naturalness: It is used to score the adversarial images for being admissible,
	    i.e. visibly not very different from the original image. We use
		Frechet Inc\'eption Distance (FID)~\cite{neuronCoverage} to measure
		naturalness. Values of FID close to 0 indicate that the
		adversarial images are natural, and are therefore desirable.

    \item Output impartiality/Pielou Score~\cite{neuronCoverage}: It reflects whether or not the adversarial image generation is biased towards any one of the output classes. It can range from 0 (biased) to 1 (unbiased).
    \begin{equation*}
	    \mathit{Pielou~Score}=\sum_{i=1}^{|\mathit{Classes}|}{\frac{\mathit{freq}_i}{\sum_{i}{\mathit{freq}_i}}*log{\frac{\mathit{freq}_i}{\sum_{i}{\mathit{freq}_i}}}}
    \end{equation*}
		$\mathit{freq}_i$=Frequency of $i^{th}$ class in the adversarial image set.
  
    \item Transferability: 
    Reflects whether the adversarial images produced by any given method are still adversarial for an adversarially trained model. We measure this by feeding the generated set of adversarial images to an adversarially trained DNN and calculate the percentage of images which are misclassified.

    A transferability of 40\% for method `A' means that 40\% of the adversarial images produced using `A' were misclassified by the Deep Neural Network which had been re-trained/hardened against adversarial attacks and the remaining 60\% were classified as their true labels.

\end{enumerate}

\begin{figure}[ht]
\centering
\captionsetup{justification=centering}
\includegraphics[width=0.9\columnwidth]{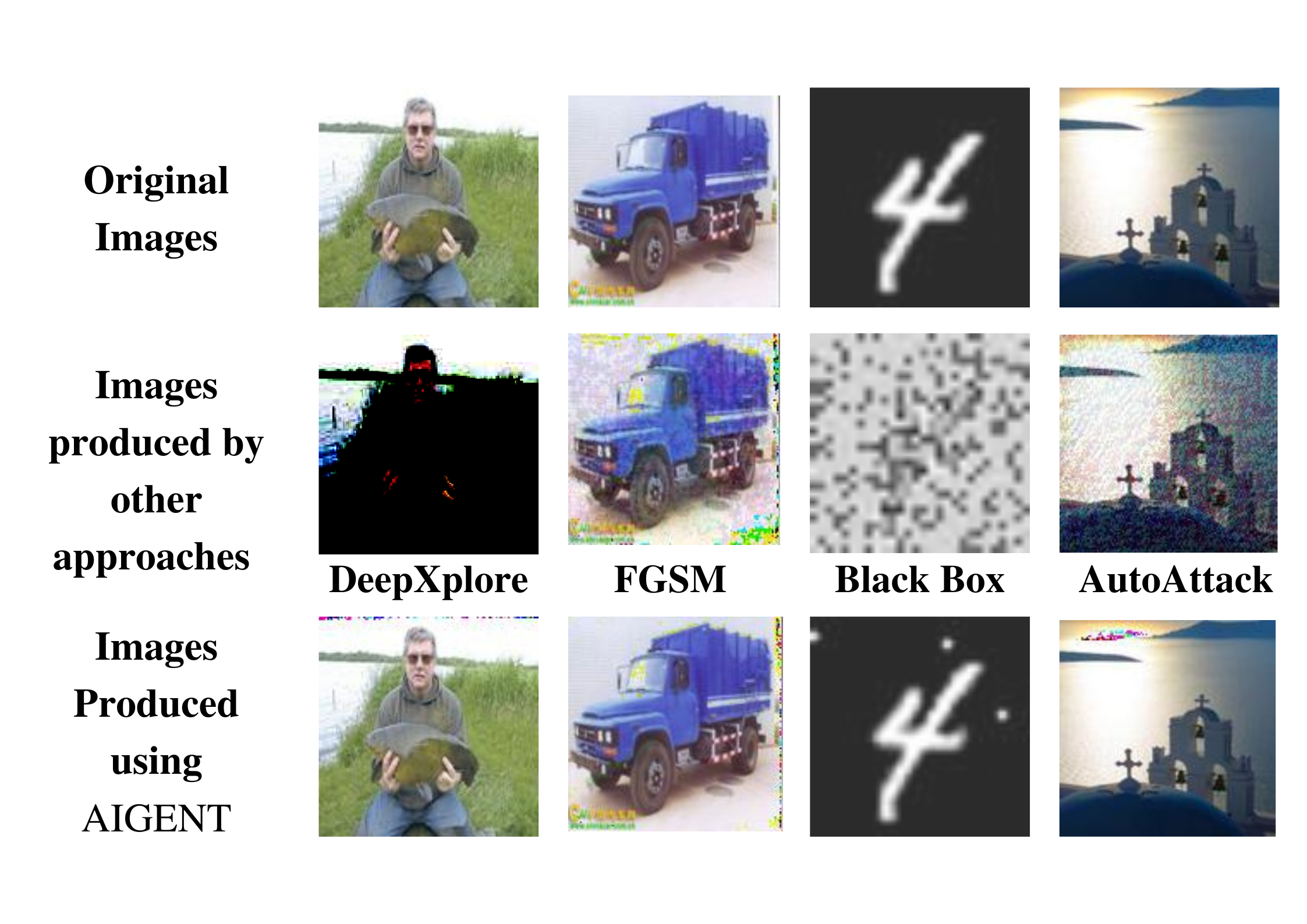} 
\caption{Examples of adversarial images generated by other techniques lacking originality.}
\label{fig17}
\end{figure}

\cite{neuronCoverage} have observed that neuron coverage was negatively
correlated with defect detection, naturalness, and output impartiality.
Naturalness is considered an essential metric while assessing the quality of
adversarial images. Fig.~\ref{fig17} shows how some of the existing methods
generate images that perform well on L2 and L-$\infty$ distance metric, \hl{but are visibly different from the original image.}

\section*{Results}
\label{sec:results}

Table~\ref{tab:2} presents a comparison of \textsc{Aigent} with other
state-of-the-art approaches of generating adversarial images, on all the three
benchmarks datasets, for several important metrics including FID, Pielou score,
defect detection rate, L-2 and L-$\infty$ distance, and the number of pixels
modified. The results demonstrate that \textsc{Aigent} performs better than all
the other approaches in terms of FID, which indicates that the adversarial
images generated by our method are natural and visibly quite similar to the
corresponding original images. This is further reinforced by the fact that
\textsc{Aigent} modifies far fewer pixels as compared to the other approaches.

\begin{table*}[htp]
\centering
\captionsetup{justification=centering}
\begin{adjustbox}{max width=\textwidth}
\begin{tabular}{| p{0.03\linewidth} | p{0.19\linewidth} | p{0.1\linewidth} | p{0.07\linewidth} | p{0.06\linewidth} | p{0.09\linewidth} | p{0.08\linewidth} | p{0.12\linewidth} | p{0.13\linewidth} | p{0.13\linewidth} |}
\hline
     
     S. No. & Technique & FID & Pielou score & L-2 & L-$\infty$ & Time ~~~~~~ (s) & Pixels modified & \% of pixels modified & Defect ~~~~~ detection \\ \hline
     \multicolumn{10}{|c|}{Benchmark dataset: MNIST}  \\\hline
	 
	  1 & \textsc{Aigent} &  \textcolor{black}{\bf 0.001} & 0.725 & \textcolor{black}{\bf 1.82} & 0.66 & 1.726 & \textcolor{black}{\bf 24} & \textcolor{black}{\bf 3.06\%} & 72.00\% \\
     
	  2 & \textsc{Aigent} (high defect)$^1$ & 0.03 & 0.74 & 4.1 & 0.80 & 1.799 & \textcolor{black}{\bf 24} & \textcolor{black}{\bf 3.06\%} & 91.40\% \\
     
     3 & FGSM$^2$ & 1.73 & \textcolor{black}{\bf 0.95} & 2.8 & \textcolor{black}{\bf 0.1} & 0.069 & 784 & 100.00\% & \textcolor{black}{\bf 99.00\%} \\
     
     4 & Black Box$^3$ & 1.98 & 0.14 & 6.58 & 0.23 & \textcolor{black}{\bf 0.065} & 784 & 100.00\% & 88.40\% \\
     
     5 & DeepXplore & 0.02 & 0.47 & 5.16 & 1 & 11.74 & 60 & 7.65\% & 45.66\% \\
     
     6 & DLFuzz$^4$ & 0.17 & 0.88 & 2.29 & 0.39 & 30 & 586 & 74.74\% & 92.36\% \\
     
     7 & AutoAttack$^5$\cite{autoattack} & 0.248 & 0.836 & 5.52 & 0.3 & 0.2 & 616 & 78.57\% & 98.40\% \\\hline
     
     
	\multicolumn{10}{|c|}{Benchmark dataset: CIFAR-10}  \\\hline
     1 & \textsc{Aigent} & \textcolor{black}{\bf 0.00009} & 0.927 & 1.6 & 0.5 & 12.01 & \textcolor{black}{\bf 12} & \textcolor{black}{\bf 0.39\%} & \textcolor{black}{\bf 100.0\%}\\
     
     2 & FGSM$^2$ & 0.071 & 0.92 & 5.5 & 0.1 & \textcolor{black}{\bf 0.079} & 3072 & 100.00\% & \textcolor{black}{\bf 100.0\%} \\
     
     3 & Black Box$^3$ & 0.44 & 0.703 & 13.04 & 0.23 & 0.082 & 3072 & 100.00\% & 76.20\% \\

     4 & AutoAttack$^5$\cite{autoattack} & 0.038 & \textcolor{black}{\bf 0.97} & \textcolor{black}{\bf 0.53} & \textcolor{black}{\bf 0.03137} & 0.2 & 588 & 57.42\% & 57.6\% \\
     
     
     5 & Output Diversification & 0.91 & 0.85 & 5.34 & 0.99 & 26.66 & 120 & 11.72\% & 100\% \\\hline

     \multicolumn{10}{|c|}{Benchmark dataset: ImageNet}  \\\hline
     1 & \textsc{Aigent} & \textcolor{black}{\bf 0.00011} & 0.75 & 6.81 & 0.73 & 35 & \textcolor{black}{\bf 300} & \textcolor{black}{\bf 0.61\%} & 98.60\%\\

     2 & FGSM$^2$ & 0.43 & \textcolor{black}{\bf 0.87} & 22 & \textcolor{black}{\bf 0.1} & 0.4 & 16384 & 100.00\% & 97.00\% \\
     
     3 & Black Box$^3$ & 0.05 & 0.8 & 52 & 0.4 & 0.3 & 16384 & 100.00\% & 90.00\% \\
     
     4 & DeepXplore & 0.032 & N.A & 58.04 & 0.51 & 84 & 15658 & 95.57\% & 59.13\% \\
     
     5 & DLFuzz$^4$ & 0.11 & N.A & 8.1 & 0.6 & 57 & 16102 & 98.28\% & 92.00\% \\

     6 & AutoAttack$^5$\cite{autoattack} & 0.045 & 0.73 & \textcolor{black}{\bf 4.68} & 0.1 & \textcolor{black}{\bf 0.147} & 6835 & 41.72\% & 42.31\%\\

     7 & Output Diversification & 0.71 & 0.83 & 25 & 0.99 & 53 & 1500 & 97.66\% & \textcolor{black}{\bf 100\%} \\\hline
     
\end{tabular}
\end{adjustbox}
	\caption{Comparison of \textsc{Aigent} with other state-of-the-art
	techniques on MNIST, CIFAR-10 and ImageNet datasets. Bold values
	indicate the best figure for each metric. DeepXplore and DLFuzz did not
	work on CIFAR-10. $^1$: Tuned to achieve higher defect detection.
	$^2$:Gradient Based technique \protect \cite{blackBox:2}. We have utilised FGSM with a maximum value of 0.1 for the L-$\infty$ norm. This has been done in order to get maximal defect detection.
	$^3$:\protect \cite{black}. $^4$: We were getting a few compilation
	errors in the DLFuzz code (\url{https://github.com/turned2670/DLFuzz})
	which we have fixed for this comparison. $^5$: We have used autoattack with limits of L-$\infty$ norm. Maximum values of L-$\infty$ norm for MNIST, CIFAR-10, and ImageNet are 0.3, 0.03, and 0.1 respectively. These are the best reported L-$\infty$ values for Autoattack. \hl{The values reported for L-2 and L-$\infty$ norms and the number of pixels modified are the maximum values obtained. The values reported for time taken are average values.}}
\label{tab:2}
\end{table*}
Our method could achieve 72\% defect detection for MNIST when constraints were
stricter. When we allowed the quality of generated adversarial images to
degrade slightly in order to achieve a higher defect detection, shown as
\textsc{Aigent} (high defect) in the table, we were able to generate adversarial
images for 91.4\% of the original images. Our method performs comparable to
white box methods. Although FGSM achieves higher defect detection,
they modify 100\% pixels which leads to visibly distinguishable images. Thus,
our method performs well in terms of defect detection, while keeping the
modification quite small. 
\hl{Defect detection using \textsc{Aigent} on CIFAR-10 was better than all the other approaches, while for ImageNet, the defect detection was lower than only Output Diversification, which modified 97.66\% pixels as compared to 0.61\% in the case of \textsc{Aigent}.}

For measuring the Pielou score, we took 50 original images of each class and
then calculated the frequency distribution on the classes of adversarial images
generated. \textsc{Aigent} was able to achieve a good Pielou score on all the
benchmark datasets. While techniques such as FGSM and Autoattack have a better
Pielou score, it comes at the expense of other metrics such as FID and the
percentage of pixels modified. Our method aims at striking a good balance between
these metrics as it is crucial for the quality of adversarial images.

\begin{figure}[ht]
\centering
\captionsetup{justification=centering}
\includegraphics[width=1\columnwidth]{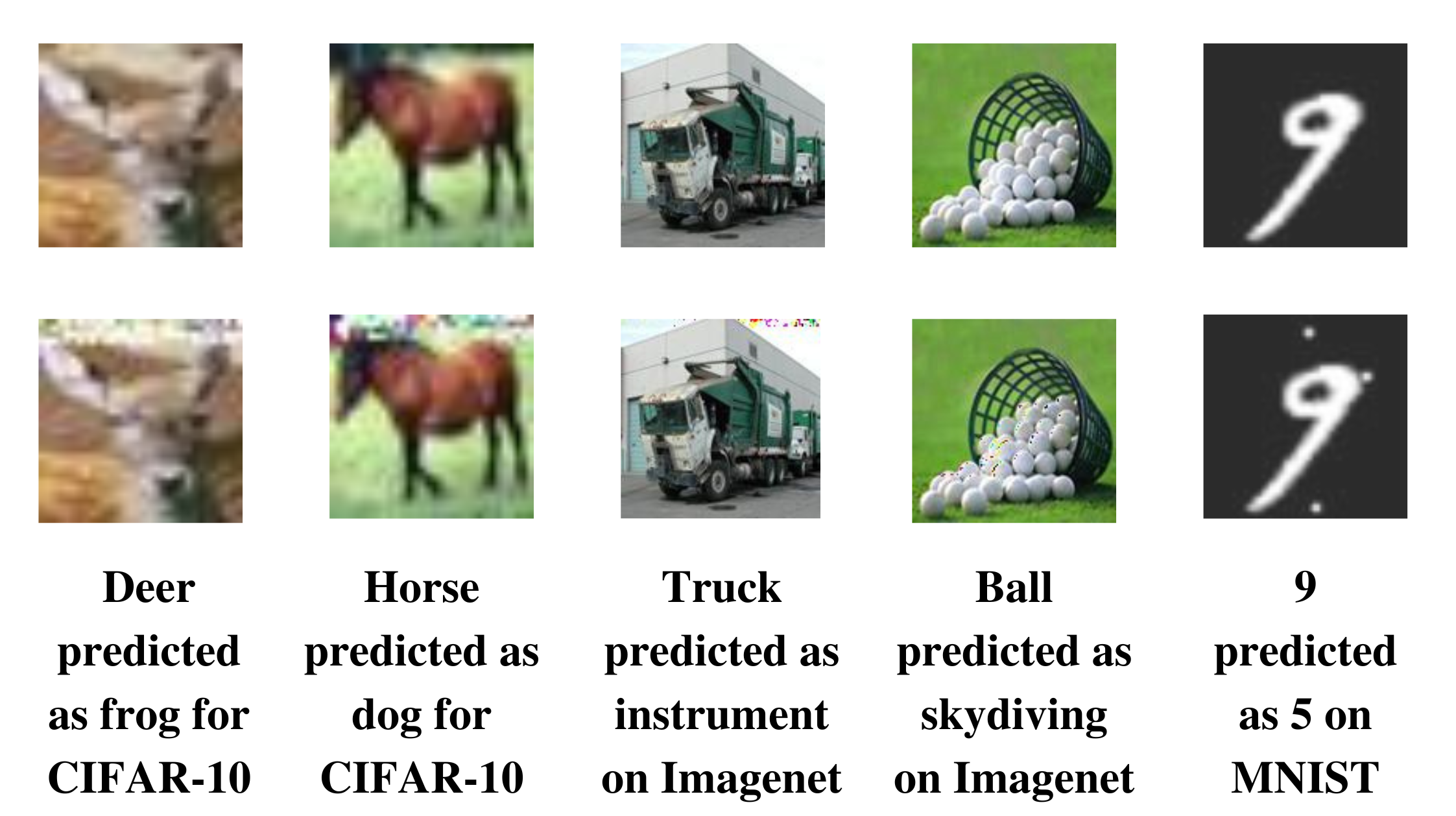} 
	\caption{Adversarial images produced by \textsc{Aigent} (bottom row), and the corresponding original images (top row).}
\label{fig8}
\end{figure}

\begin{figure}[ht]
\centering
\captionsetup{justification=centering}
\includegraphics[width=0.8\columnwidth]{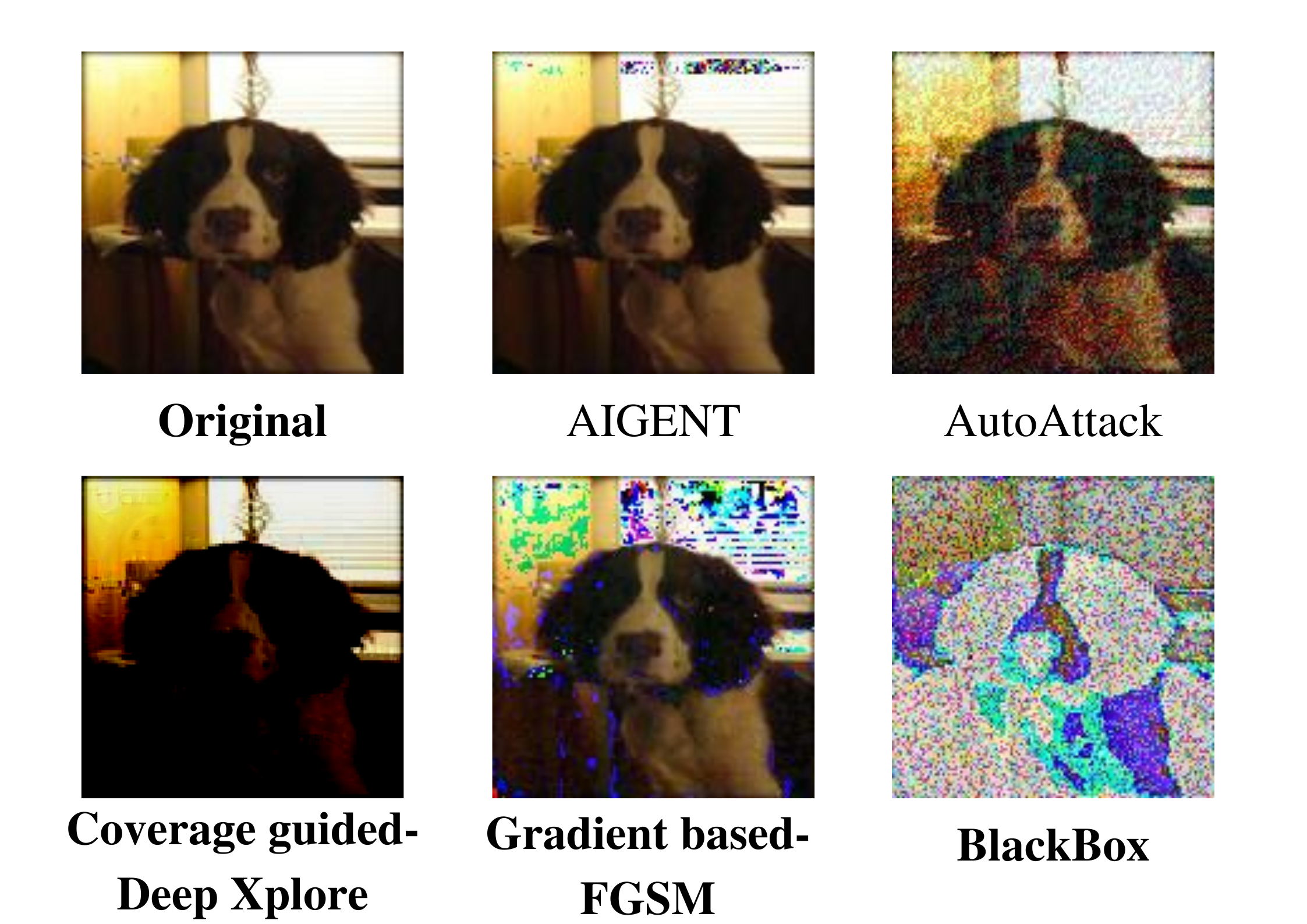} 
\caption{Sample adversarial images from different tools.}
\label{fig18}
\end{figure}

Fig.~\ref{fig8} shows sample adversarial images produced for MNIST, CIFAR-10
and ImageNet datasets. The first row contains original images and the second
row contains their corresponding adversarial images. Fig.~\ref{fig18} shows a
comparison of \textsc{Aigent} with Autoattack, DeepXplore, FGSM , and black-box, on the same input image.

\subsection*{Transferability}
We generated adversarial images using the strategies listed in Table \ref{tab:2} and randomly selected a set of adversarial images so that images generated by each technique were uniformly present in the set. Then, we re-trained deep neural network using the selected set of adversarial images to get a more robust neural network N'. For each benchmark and method listed in Table \ref{tab:2}, we then attacked the network N' and calculated the transferability score for each case. We noticed that the transferability score for \textsc{Aigent} was better than all the other methods(as shown in Table~\ref{tab:tab4}. This shows that \textsc{Aigent} produces better quality images which are able to fool hardened DNNs.

\begin{table*}[h]

\centering
\begin{adjustbox}{max width=\textwidth}
\begin{tabular}{| p{0.15\linewidth} | p{0.14\linewidth} | p{0.14\linewidth} | p{0.14\linewidth} | p{0.14\linewidth} | p{0.14\linewidth} | p{0.14\linewidth} | }
\hline & \textsc{Aigent} & FGSM & BlackBox & DLFuzz & DeepXplore & AutoAttack \\ \hline
MNIST    & 48\% & 38\% & 42\% & 32\% & 45\% & 34\%\\ \hline
CIFAR-10 & 51\% & 41\% & 43\% & 30\% & 46\% & 33\%\\ \hline
ImageNet & 42\% & 29\% & 27\% & 33\% & 37\% & 23\%\\ \hline
\end{tabular}
\end{adjustbox}
\caption{Comparison of Transferability scores for different techniques.}
\label{tab:tab4}
\end{table*}

\subsection*{Comparisions with Goldberger et al. \protect\cite{modifications:1}}
\label{sec:comp}

\begin{table*}[h]
\centering
\captionsetup{justification=centering}
\begin{adjustbox}{max width=\textwidth}
\begin{tabular}{|c|c|rr|rr|rr|rr|}
\hline
\multicolumn{1}{|l|}{} & \multicolumn{1}{l|}{} & \multicolumn{2}{l|}{\textbf{Network 1}}  & \multicolumn{2}{l|}{\textbf{Network 2}}  & \multicolumn{2}{l|}{\textbf{Network 3}} & \multicolumn{2}{l|}{\textbf{Network 4}}  \\ \hline

\textbf{Technique} & \textbf{Layer} & \multicolumn{1}{c|}{\textbf{n}} & \multicolumn{1}{c|}{\textbf{mod}} & \multicolumn{1}{c|}{\textbf{n}} & \multicolumn{1}{c|}{\textbf{mod}} & \multicolumn{1}{c|}{\textbf{n}} & \multicolumn{1}{c|}{\textbf{mod}} & \multicolumn{1}{c|}{\textbf{n}} & \multicolumn{1}{c|}{\textbf{mod}} \\ \hline

\textbf{Using AIGENT} & \textbf{0} & \multicolumn{1}{r|}{$\infty$} & $\infty$ & \multicolumn{1}{r|}{26} & 19.3277 & \multicolumn{1}{r|}{7} & 0.0653 & \multicolumn{1}{r|}{6} & 0.2092                            \\ \cline{2-10} 

& \textbf{1} & \multicolumn{1}{r|}{$\infty$} & $\infty$ & \multicolumn{1}{r|}{$\infty$} & $\infty$  & \multicolumn{1}{r|}{231} & 0.0266 & \multicolumn{1}{r|}{253} & 0.3156\\ \cline{2-10} 

& \textbf{2} & \multicolumn{1}{r|}{$\infty$} & $\infty$ & \multicolumn{1}{r|}{408} & 3.45949 & \multicolumn{1}{r|}{165} & 0.0595 & \multicolumn{1}{r|}{143} & 0.0991 \\ \cline{2-10} 

& \textbf{3} & \multicolumn{1}{r|}{$\infty$} & $\infty$  & \multicolumn{1}{r|}{336} & 5.8526 & \multicolumn{1}{r|}{255} & 0.2676 & \multicolumn{1}{r|}{299} & 0.16040 \\ \cline{2-10} 

& \textbf{4} & \multicolumn{1}{r|}{$\infty$} & $\infty$ & \multicolumn{1}{r|}{154} & 2.0554 & \multicolumn{1}{r|}{187} & 0.0620 & \multicolumn{1}{r|}{230} & 0.74573 \\ \cline{2-10} 

& \textbf{5} & \multicolumn{1}{r|}{$\infty$} & $\infty$ & \multicolumn{1}{r|}{33} & 0.3085 & \multicolumn{1}{r|}{66} & 0.00092 & \multicolumn{1}{r|}{50} & 0.00587 \\ \cline{2-10} 

& \textbf{6} & \multicolumn{1}{r|}{5} & \textcolor{blue}{\textbf{0.03364}} & \multicolumn{1}{r|}{15} & \textcolor{blue}{\textbf{0.0383}} & \multicolumn{1}{r|}{30} & \textcolor{blue}{\textbf{0.00017}} & \multicolumn{1}{r|}{15} & \textcolor{blue}{\textbf{0.00070}}  \\ \cline{2-10} 

& \textbf{7} & \multicolumn{1}{r|}{1} & 0.03488 & \multicolumn{1}{r|}{4} & 0.04002 & \multicolumn{1}{r|}{3} & 0.00019 & \multicolumn{1}{r|}{5} & \textcolor{blue}{\textbf{0.00070}} \\ \hline

\textbf{Using Goldberber et al} & \textbf{7} & \multicolumn{1}{r|}{7} & 3.1 & \multicolumn{1}{r|}{5} & 0.08 & \multicolumn{1}{r|}{2} & 0.003 & \multicolumn{1}{r|}{2} & 0.004 \\ \hline

\end{tabular}
\end{adjustbox}
	\caption{Comparison of modifications found by \textsc{Aigent} and \protect\cite{modifications:1} in different layers of the given neural networks. [n=number of weights modified; mod=Total modification in the n weights; $\infty$=Infeasible]. Networks 1-4 are ACASXu Networks. The values in bold (and blue) indicate the least modifications.}
\label{tab:3}
\end{table*}

Our tool \textsc{Aigent} implements an improvement over the modification technique (explained in section 4.1 of the paper) used in \cite{modifications:1}. To quantify the benefits of this improvement, we have compared our network patching technique with the technique proposed in \cite{modifications:1} using the same setup used in \cite{modifications:1}.

Table ~\ref{tab:3} shows the modifications found using our network patching technique and the technique mentioned in \cite{modifications:1}. We note that the modifications found using our technique are invariably smaller than the ones found by \cite{modifications:1}. The tool given by \cite{modifications:1} finds modifications only in the last layer of the network (the technique, however, does not have this limitation). \textsc{Aigent} not only finds smaller modifications but it is also faster. The time taken by \textsc{Aigent} to find last layer modifications was only 3 seconds, while \cite{modifications:1} took 30 seconds to find modifications for the same objective function.

\subsection*{Overcoming a Methodological Limitation: Proposed Approach and Empirical Findings}
It is noteworthy that in Algorithm \ref{alg:alg1}, while finding modifications in a particular layer, we assign each neuron to a particular phase. For this, we first generate the value of every neuron for the given input and use that to assign a phase to the neurons. For example, if for input $i$ the $1^{st}$ neuron in Layer 2 had a positive value, we would fix the phase of that neuron to \emph{active} (i.e., $x\ge0$, and thus $ReLU(x)=x$), or else we will fix the phase to \emph{inactive} (i.e., $x<0$, and thus $ReLU(x)=0$). In our algorithm, active phase neurons are only allowed to increase and inactive phase neurons are only allowed to decrease.

While our technique works well in practice and finds an adversarial example in every case, it may fail in doing so if the assigned phases do not contain an adversarial example. Even when we find an adversarial example, it may happen that a different phase-combination leads to a ``better'' adversarial example. Checking all phase-combinations, however, will lead to an exponential number of calls to \textsc{Aigent}. To overcome this, we used linear approximations for all the neurons in the network instead of fixing their phases. We first calculated linear approximations for all activation functions in the network using the technique mentioned in \cite{approximations}. We calculated approximations for a particular class, which means that for a set of $\mathcal{K}$ classes, we generated $\mathcal{K}$ sets of linear approximations. Thus, for a total of $\mathcal{R}$ number of ReLU neurons in the network and $\mathcal{K}$ classes, we will have $\mathcal{K*R}$ linear approximations. Since these approximations are calculated for the entire dataset as a pre-processing step, adversarial examples can be generated faster. We observed that using linear approximations with \textsc{Aigent} decreased the average time taken\footnote{The average time taken includes the time taken to compute the approximations.} to generate adversarial images without compromising on the quality (FID, L-2, and L-$\infty$ norm) of the images generated. The times decreased from 1.7, 12.01, and 35 seconds to 1.5, 10.3, and 31 seconds for MNIST, CIFAR-10, and ImageNet, respectively. 

\section*{Conclusion}
\label{sec:conc}

Finding adversarial inputs for DNNs is not just useful for identifying
situations when a network may behave unexpectedly, but also for adversarial
training, which can make the network robust. We have proposed a technique to
generate adversarial inputs via patching of neural networks. In our experiments
over three benchmark image datasets, we observed that the proposed method is
significantly more effective than the existing state-of-the-art -- it could
generate natural adversarial images (FID scores $\leq 10^{-3}$) by perturbing a
tiny fraction of pixels ($\approx$ 3\% in the worst case).

There are several interesting directions for future work.  Since the proposed
method works by finding a patch repeatedly, better algorithms for DNN patching
would also make our technique more efficient.  Another useful direction would
be to find a \emph{minimal} patch in order to get the \emph{closest}
adversarial example. And even though the technique can, in principle, be
applied to DNNs with any activation function, it would be worthwhile to
engineer our approach to handle activations other than ReLU efficiently.

\bibliographystyle{IEEEtran}
\bibliography{IEEEabrv,prdc}

\end{document}